\documentclass[11pt,a4paper]{article}
\usepackage[utf8]{inputenc}
\usepackage{amsmath}
\usepackage{amsfonts}
\usepackage{amssymb}
\usepackage{graphicx}
\usepackage{booktabs} 
\usepackage{xcolor}   
\usepackage{xurl} 
\usepackage{listings} 
\usepackage[numbers,sort&compress]{natbib} 
\usepackage[breaklinks=true]{hyperref} 
\usepackage{authblk} 
\usepackage{siunitx} 
\usepackage{booktabs}   
\usepackage{tabularx}   
\usepackage{ragged2e}   

\title{Two-Stage Reasoning-Infused Learning: Improving Classification with LLM-Generated Reasoning}

\author[1]{Mads Henrichsen}
\author[2]{Rasmus Krebs}
\affil[1]{syv.ai}
\affil[2]{syv.ai}

\date{\today}

\begin{document}
\maketitle

\begin{abstract}
Standard classification models often map inputs directly to labels without explicit reasoning, potentially limiting their performance, robustness, and interpretability. This paper introduces a novel two-stage approach to enhance text classification by leveraging Large Language Model (LLM)-generated reasonings. In the first stage, we fine-tune a Llama-3.2-1B-Instruct model (henceforth \texttt{Llama-R-Gen}) on a general-purpose reasoning dataset (\texttt{syvai/reasoning-gen}) to generate textual reasoning (R) given a question and its answer. In the second stage, this generally trained \texttt{Llama-R-Gen} is used offline to create an augmented training dataset for a downstream generative model. This downstream model, based on Llama-3.2-1B-Instruct, takes only the input text (Q) and is trained to output the generated reasoning (R) immediately followed by the predicted emotion (A). We demonstrate this methodology on the \href{https://huggingface.co/datasets/dair-ai/emotion}{dair-ai/emotion} dataset for emotion classification. Our experiments show that the generative model trained to output reasoning and the emotion (\texttt{Classifier\_Q->RA}) achieves a significant improvement of \textbf{8.7 percentage points} in accuracy (for emotion prediction) compared to a baseline generative model trained solely to output the emotion (\texttt{Classifier\_Q->A}), highlighting the strong generalization capabilities of the reasoning generation and the benefit of explicit reasoning training. This work underscores the potential of LLM-generated reasonings for creating richer training datasets, thereby improving the performance of diverse downstream NLP tasks and providing explicit explanations.
\end{abstract}

\pagebreak

\section{Introduction}
Text classification systems are pivotal in numerous applications, from sentiment analysis to spam detection, aiming to categorize text into predefined labels \cite{jurafsky2000speech}. However, many contemporary text classification models operate as "black boxes," directly mapping text to labels without an explicit intermediate reasoning process \cite{ribeiro2016should}. This lack of transparency can hinder model performance, particularly on complex inputs requiring multi-step inference or nuanced understanding, and makes it difficult to diagnose failures or build trust in the system's outputs.

The ability to reason is a hallmark of human intelligence and is increasingly recognized as a crucial component for advancing artificial intelligence \cite{wei2022chain}. While large language models (LLMs) have shown impressive capabilities in generating coherent text and performing in-context learning \cite{brown2020language, touvron2023llama, ai2024meta}, explicitly incorporating reasoning into the training paradigm of downstream task models remains an active area of research.

In this paper, we propose a two-stage framework to improve text classification performance by infusing training data with LLM-generated reasonings. Our core hypothesis is that training a model to explicitly generate a reasoning path alongside its prediction will enable it to learn more robust representations and make more accurate predictions for the target label. Furthermore, by directly generating this reasoning, our system inherently provides both the predicted label and the explanatory reasoning behind it. While the framework is generally applicable to various text classification tasks, we demonstrate its efficacy on emotion classification using the \href{https://huggingface.co/datasets/dair-ai/emotion}{dair-ai/emotion} dataset.

The two stages are:
\begin{enumerate}
    \item \textbf{Reasoning Generation (Llama-R-Gen):} We fine-tune a Llama-3.2-1B-Instruct model (\texttt{Llama-R-Gen}) on a general-purpose reasoning dataset. This training teaches the model to generate step-by-step reasoning given a question and its corresponding answer.
    \item \textbf{Reasoning-Generated Classification (\texttt{Classifier\_Q->RA}):} The \texttt{Llama-R-Gen} model is then used offline to create an augmented training dataset for a downstream generative model. This downstream model, based on Llama-3.2-1B-Instruct, is trained to take only the input text (Q) and directly generate a combined sequence of the reasoning (R) and the predicted class (A). This integrated generation ensures that the model outputs both the reasoning and the predicted emotion as a single coherent response.
\end{enumerate}

Our contributions are threefold:
\begin{itemize}
    \item We present a methodology for fine-tuning a model on a general reasoning dataset to develop a reasoning generation model, designed to be transferable to new domains.
    \item We introduce a novel dataset augmentation strategy that enriches text classification datasets by creating (Text, Reasoning + Label) pairs, training a downstream generative model to produce explicit reasonings alongside its predictions.
    \item We provide a comprehensive validation of our methodology on the \href{https://huggingface.co/datasets/dair-ai/emotion}{dair-ai/emotion} dataset, demonstrating that our reasoning-infused learning approach significantly improves emotion classification accuracy compared to a strong baseline.
\end{itemize}

The remainder of this paper is structured as follows: Section \ref{sec:related_work} discusses related work. Section \ref{sec:methodology} details our two-stage methodology. Section \ref{sec:experiments} describes the experimental setup, datasets, and evaluation metrics. Section \ref{sec:results} presents and analyzes the results. Section \ref{sec:discussion} discusses the implications and limitations of our findings, and Section \ref{sec:conclusion} concludes the paper.

\section{Related Work}
\label{sec:related_work}
Our work builds upon several key areas in natural language processing.

\paragraph{Chain-of-Thought Prompting.} The popularization of Chain-of-Thought (CoT) prompting has shown that eliciting intermediate reasoning steps from LLMs at inference time can significantly improve performance on complex tasks \cite{wei2022chain, kojima2022large}. These methods typically apply to very large, general-purpose models in a few-shot or zero-shot setting. In contrast, our work adapts this principle for smaller, fine-tuned models. Instead of prompting for a reasoning path at inference, we pre-generate reasonings to create a richer training dataset, aiming to distill the reasoning capability into a more efficient, task-specific model.

\paragraph{Explainable AI (XAI) in NLP.} A growing body of research aims to make NLP models more transparent by generating explanations for their predictions. Some approaches train models to extract text snippets as rationales \cite{lei2016rationalizing}, while others use human-annotated explanations for supervision, as seen in datasets like e-SNLI \cite{camburu2018snli}. For example, \citet{rajani2019explain} demonstrated that training on human-written explanations can improve model performance and generalization. Our work aligns with this goal but differs in methodology: we use a general-purpose LLM to *generate* explanations automatically, thereby reducing the dependency on costly human annotation and enabling scalability to datasets without existing explanations.

\paragraph{Dataset Augmentation.} Data augmentation is a standard technique for improving model generalization by increasing training data size and diversity \cite{shorten2019survey}. Traditional NLP methods include back-translation or synonym replacement. Our approach introduces a novel form of augmentation by synthesizing structured, explanatory content (the reasoning) and prepending it to the target label. This provides a much richer supervisory signal than simple label-to-text mapping, pushing the model to learn the "why" behind a prediction, not just the "what."

\paragraph{Learning with Reasonings.} Prior work has explored jointly training models to predict labels and generate explanations. \citet{wiegreffe2021teach} explored various settings for "learning from explanations," showing that explanations can serve as a valuable supervisory signal. Our two-stage framework provides a practical and scalable method to achieve this. By first training a dedicated reasoning generator on a broad corpus and then using it to augment data for a separate downstream classifier, we decouple the complex task of general reasoning from the specific classification task, allowing each model to specialize. This modular approach contrasts with end-to-end systems and proves effective for transferring reasoning skills to a new domain.

\section{Methodology}
\label{sec:methodology}
\begin{figure}[!ht]
    \centering
    \includegraphics[width=0.9\textwidth]{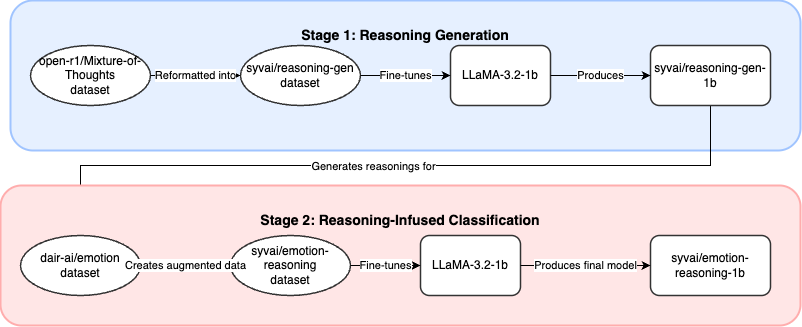} 
    \caption{Overview of the two-stage reasoning-infused learning framework. Stage 1 involves fine-tuning \texttt{Llama-R-Gen} on a general dataset to learn how to generate reasoning (R) from (Question, Answer) pairs. Stage 2 uses the trained \texttt{Llama-R-Gen} to create an augmented dataset for a downstream task. This dataset is then used to fine-tune a generative classifier, which learns to predict the emotion (A) by generating the reasoning (R) first, based only on the input text.}
    \label{fig:framework}
\end{figure}

Our proposed two-stage reasoning-infused learning framework is depicted in Figure \ref{fig:framework}. We first train a reasoning generation model and then use its output to construct an augmented training dataset for a downstream generative classifier.

\subsection{Stage 1: Reasoning Generation Model (\texttt{Llama-R-Gen})}
The goal of this stage is to develop a general-purpose model capable of generating a plausible textual reasoning (R) given an input question (Q) and its correct answer (A).

\paragraph{Model Architecture.} We utilize Llama-3.2-1B-Instruct, a decoder-only transformer model with 1 billion parameters, known for its strong generative capabilities \cite{ai2024meta}.

\paragraph{Training Data for Reasoning Generation.}
\label{sec:rationale_data_creation}
To fine-tune \texttt{Llama-R-Gen}, we utilize the \href{https://huggingface.co/datasets/syvai/reasoning-gen}{\texttt{syvai/reasoning-gen}} dataset. This dataset is derived from the `open-r1/Mixture-of-Thoughts` dataset. The key transformation was to restructure the original data, which contained multi-turn conversational thoughts, into a direct `(Question, Answer) -> Reasoning` format. This was done to explicitly teach a model to generate a complete reasoning process when provided with a problem and its solution.
The dataset contains approximately \textbf{350,000} such triples across diverse domains like math, code, and science. This general-purpose dataset is crucial for our goal of training a model that can generalize its reasoning ability to new domains like emotion classification. We used an 80/20 split for training/validation.

\paragraph{Fine-tuning Process.}
\texttt{Llama-R-Gen} was fine-tuned on the \texttt{syvai/reasoning-gen} dataset. The input to the model was formatted as a single sequence:
\texttt{"Question: [Q\_text] Answer: [A\_text] Reasoning: "}
The model was trained to predict the gold reasoning $R_{gold}$ using a standard language modeling objective (cross-entropy loss). Key fine-tuning hyperparameters are detailed in Appendix \ref{app:hyperparams}.

\subsection{Stage 2: Reasoning-Generated Emotion Classification}
In this stage, we first use the generally trained \texttt{Llama-R-Gen} to create augmented training data for our downstream generative classifier models.

\paragraph{Base Classification Dataset ($D_{target}$).}
We use the \href{https://huggingface.co/datasets/dair-ai/emotion}{dair-ai/emotion} dataset as our target task. This dataset consists of text inputs labeled with one of 6 basic emotions: sadness, joy, love, anger, fear, and surprise. Table \ref{tab:dataset_distribution} shows the class distribution of the test set, highlighting its imbalanced nature.

\begin{table}[h!]
\centering
\caption{Class distribution of the \href{https://huggingface.co/datasets/dair-ai/emotion}{dair-ai/emotion} test set (N=2000).}
\label{tab:dataset_distribution}
\begin{tabular}{lcc}
\toprule
\textbf{Emotion} & \textbf{Count} & \textbf{Percentage (\%)} \\
\midrule
Joy & 695 & 34.8 \\
Sadness & 581 & 29.1 \\
Anger & 275 & 13.8 \\
Fear & 224 & 11.2 \\
Love & 159 & 8.0 \\
Surprise & 66 & 3.3 \\
\bottomrule
\end{tabular}
\end{table}

\paragraph{Dataset Augmentation for Generative Classifiers.}
For each instance $(Q_i, A_{i,correct})$ in the training split of `dair-ai/emotion`, we generate a reasoning $R_i$ using the fine-tuned \texttt{Llama-R-Gen} model. The input prompt for this step is: \texttt{"Question: [$Q_i$] Answer: [$A_{i,correct}$] Reasoning: "}. This process is performed once offline to construct the training data.

The target output sequences for our models are then constructed:
\begin{itemize}
    \item For our \textbf{proposed} model (\texttt{Classifier\_Q->RA}), the target is $T_{i,RA} = R_i + \text{" "} + A_{i,correct}$.
    \item For our \textbf{baseline} model (\texttt{Classifier\_Q->A}), the target is simply $T_{i,A} = A_{i,correct}$.
\end{itemize}
This results in two datasets made publicly available: \href{https://huggingface.co/datasets/syvai/emotion-reasoning}{\texttt{syvai/emotion-reasoning}} for the proposed model and \href{https://huggingface.co/datasets/syvai/no-emotion-reasoning}{\texttt{syvai/no-emotion-reasoning}} for the baseline.

\paragraph{Downstream Generative Classifier Models.}
Both the proposed (\texttt{Classifier\_Q->RA}) and baseline (\texttt{Classifier\_Q->A}) models are fine-tuned from \textbf{Llama-3.2-1B-Instruct}. Both models take only the original text $Q_i$ as input, prompted as follows:
"Find the emotion in the text." (system message)
\texttt{"[$Q_i$]"} (user message)

The models are then trained to generate their respective target sequences ($T_{i,RA}$ or $T_{i,A}$) using a standard cross-entropy loss. Key fine-tuning hyperparameters are detailed in Appendix \ref{app:hyperparams}.

\paragraph{Inference Workflow.}
At inference time, a user provides a text (Q). The fine-tuned \texttt{Classifier\_Q->RA} model takes this text as input and directly generates a single sequence containing both the reasoning (R) and the predicted emotion (A), providing an interpretable output.

\section{Experiments}
\label{sec:experiments}
\subsection{Datasets}
\begin{itemize}
    \item \textbf{\texttt{D\_reasoning\_seed}}: The \href{https://huggingface.co/datasets/syvai/reasoning-gen}{\texttt{syvai/reasoning-gen}} dataset (~350k examples) was used to train \texttt{Llama-R-Gen}.
    \item \textbf{$D_{target}$ (\href{https://huggingface.co/datasets/dair-ai/emotion}{dair-ai/emotion})}: Official splits were used: 16,000 training, 2,000 validation, and 2,000 test instances.
\end{itemize}

\subsection{Models and Baselines}
\begin{itemize}
    \item \textbf{Proposed Generative Classifier (\texttt{Classifier\_Q->RA})}: Llama-3.2-1B-Instruct fine-tuned on the reasoning-augmented `syvai/emotion-reasoning` dataset.
    \item \textbf{Baseline Generative Classifier (\texttt{Classifier\_Q->A})}: Llama-3.2-1B-Instruct fine-tuned on `syvai/no-emotion-reasoning`, predicting only the emotion.
    \item \textbf{GPT-4.1 (Zero-Shot Baseline)}: A powerful general-purpose LLM used for a zero-shot performance benchmark without any task-specific fine-tuning.
\end{itemize}

\subsection{Experimental Setup}
All models were fine-tuned using the \textbf{Axolotl} framework on an \textbf{NVIDIA A40 GPU}. The generation of reasonings for the data augmentation step was performed using the vLLM inference engine.

\subsection{Evaluation Metrics}
The primary metric is \textbf{Accuracy} for the predicted emotion label. We also report per-class \textbf{Precision, Recall, and F1-score}, along with their macro and weighted averages.

\section{Results}
\label{sec:results}
\subsection{Qualitative Analysis of Generated Reasonings}
Table \ref{tab:rationale_examples} presents a qualitative analysis of outputs from our proposed \texttt{Classifier\_Q->RA} model, illustrating different success and failure modes. The model can produce coherent reasonings that correctly justify its predictions, but it can also make errors where the reasoning is flawed or misaligned with the text.

\begin{table}[h!]
\centering
\caption{Qualitative examples of outputs from the proposed \texttt{Classifier\_Q->RA} model on the test set.}
\label{tab:rationale_examples}
\resizebox{\textwidth}{!}{%
\begin{tabular}{p{0.3\linewidth} p{0.15\linewidth} p{0.15\linewidth} p{0.35\linewidth}}
\toprule
\textbf{Text} & \textbf{True Label} & \textbf{Predicted Label} & \textbf{Generated Reasoning} \\
\midrule
\multicolumn{4}{c}{\textbf{Case 1: Correct Prediction with Good Reasoning}} \\
\midrule
i am feeling very sad and sorry for my self that i am not at home with my family on this day of eid & sadness & sadness & The text explicitly mentions feeling "very sad and sorry," which are clear indicators of the sadness emotion. The context of being away from family on a special day reinforces this feeling. \\
\midrule
\multicolumn{4}{c}{\textbf{Case 2: Correct Prediction with Flawed Reasoning}} \\
\midrule
i feel like i am a strange little doll that has been wound up and is running about but with no real feelings & fear & fear & The text describes a feeling of detachment and being controlled, like a "strange little doll." This sense of alienation and lack of agency points toward joy. \\
\midrule
\multicolumn{4}{c}{\textbf{Case 3: Incorrect Prediction with Plausible Reasoning}} \\
\midrule
i feel that i am useful to my team and i am a good contributor & joy & love & The user expresses feelings of being "useful" and a "good contributor" to their team. This strong sense of belonging, appreciation, and positive connection to a group aligns with the emotion of love. \\
\midrule
\multicolumn{4}{c}{\textbf{Case 4: Incorrect Prediction with Flawed Reasoning}} \\
\midrule
i feel a little shaky and insecure & fear & joy & The text mentions feeling "a little shaky." This could be interpreted as a physical reaction to a positive event, like excitement or exhilaration, which are associated with joy. \\
\bottomrule
\end{tabular}
}
\end{table}

\subsection{Downstream Emotion Classification Performance}
The main results comparing our proposed reasoning-augmented classifier with the baselines are presented in Table \ref{tab:main_results}. Our proposed method, \texttt{Classifier\_Q->RA}, achieves an accuracy of \textbf{58.4\%}. This result significantly outperforms the fine-tuned baseline (\texttt{Classifier\_Q->A} at 49.7\%) by 8.7 absolute percentage points. A two-proportion z-test confirms that this improvement is statistically significant ($z = 6.88, p < .001$). Furthermore, our model surpasses the powerful GPT-4.1 zero-shot baseline by 26.4 percentage points, highlighting the effectiveness of specialized, reasoning-infused fine-tuning.

\begin{table}[h!]
\centering
\caption{Emotion classification accuracy on the \href{https://huggingface.co/datasets/dair-ai/emotion}{dair-ai/emotion} test set.}
\label{tab:main_results}
\begin{tabular}{lc}
\toprule
\textbf{Model} & \textbf{Accuracy (\%)} \\
\midrule
GPT-4.1 (Zero-Shot Baseline) & 32.0 \\
\texttt{Classifier\_Q->A} (Fine-tuned Baseline) & 49.7 \\
\textbf{\texttt{Classifier\_Q->RA} (Proposed)} & \textbf{58.4} \\
\midrule
Improvement (Proposed vs. Fine-tuned Baseline) & \textbf{+8.7} \\
\bottomrule
\end{tabular}
\end{table}

\subsubsection{Per-Emotion Accuracy and F1-Scores}
Table \ref{tab:per_emotion_accuracy} and Table \ref{tab:f1_scores} provide a detailed breakdown of performance. The \texttt{Classifier\_Q->RA} model shows substantial gains over the baseline for several key emotions: sadness (+19.6\%), anger (+4.0\%), and fear (+18.2\%). However, performance for the "surprise" class dropped significantly. This suggests that while the reasoning augmentation was highly beneficial for common classes, it may have been detrimental for the severely underrepresented "surprise" class. The macro and weighted F1-scores further confirm the overall superiority of the proposed model, indicating a better-balanced and more robust classifier.

\begin{table}[h!]
\centering
\caption{Per-Emotion Accuracy for All Classifiers (\%).}
\label{tab:per_emotion_accuracy}
\begin{tabularx}{0.95\linewidth}{l >{\Centering}X >{\Centering}X >{\Centering\bfseries}X}
\toprule
\textbf{Emotion} & \textbf{GPT-4.1 (Zero-Shot)} & \textbf{\texttt{Classifier\_Q->A} (Baseline)} & \textbf{\texttt{Classifier\_Q->RA} (Proposed)} \\
\midrule
Sadness & 27.9 & 44.3 & \textbf{63.9} \\
Joy & 52.2 & 73.5 & \textbf{75.5} \\
Love & 12.5 & \textbf{21.0} & 20.8 \\
Anger & 33.3 & 40.7 & \textbf{44.7} \\
Fear & 20.4 & 32.7 & \textbf{50.9} \\
Surprise & 2.9 & \textbf{13.8} & 1.5 \\
\bottomrule
\end{tabularx}
\end{table}

\begin{table}[h!]
\centering
\caption{Macro and Weighted Average F1-Scores for All Classifiers.}
\label{tab:f1_scores}
\begin{tabularx}{0.95\linewidth}{l >{\Centering}X >{\Centering}X >{\Centering\bfseries}X}
\toprule
\textbf{Metric} & \textbf{GPT-4.1 (Zero-Shot)} & \textbf{\texttt{Classifier\_Q->A} (Baseline)} & \textbf{\texttt{Classifier\_Q->RA} (Proposed)} \\
\midrule
Macro Avg F1 & 0.2500 & 0.3975 & \textbf{0.4317} \\
Weighted Avg F1 & 0.3200 & 0.4923 & \textbf{0.5695} \\
\bottomrule
\end{tabularx}
\end{table}

\begin{figure}[htbp]
\centering
\includegraphics[width=0.9\textwidth]{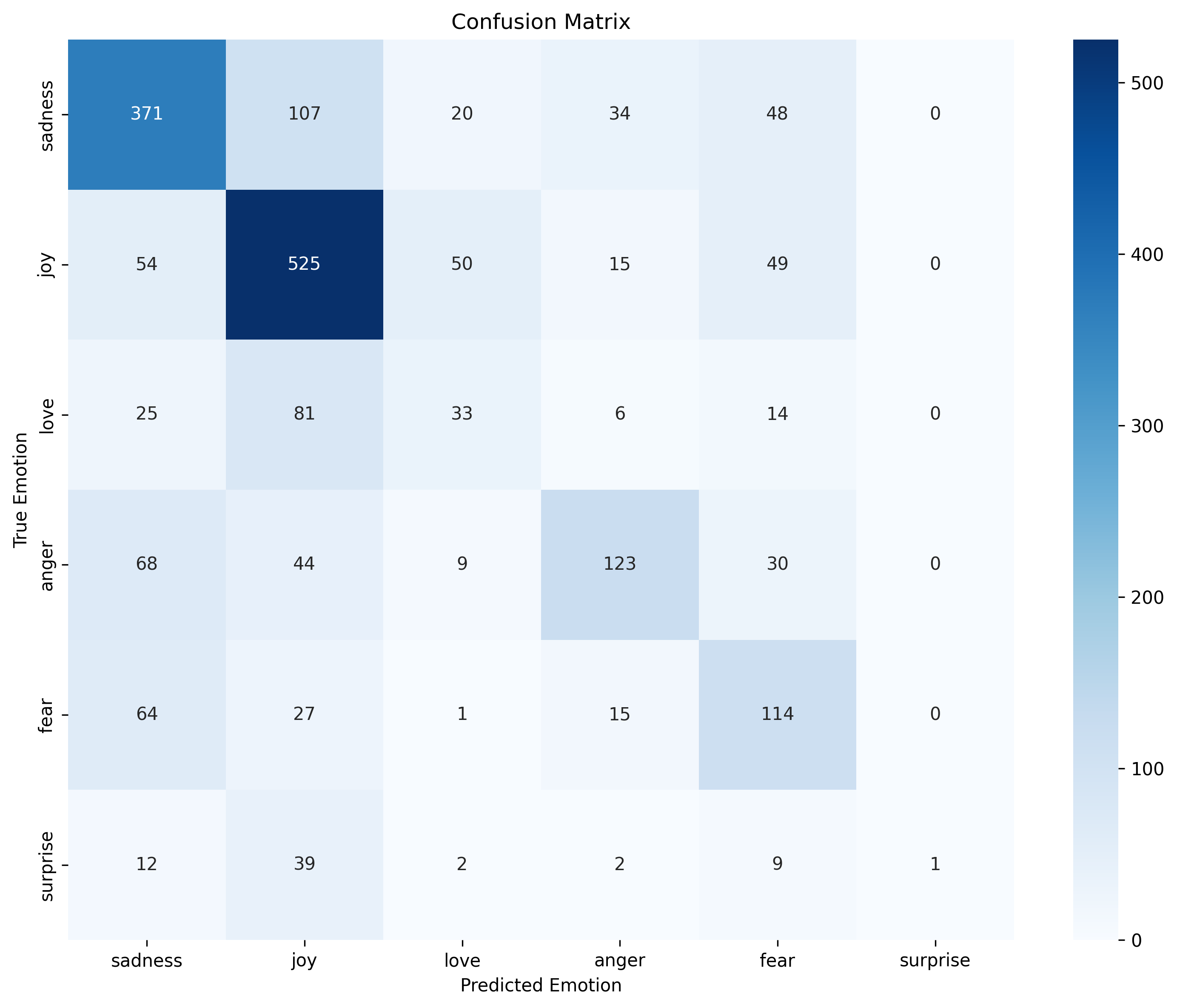}
\caption{Baseline Classifier (\texttt{Classifier\_Q->A}): Confusion Matrix.}
\label{fig:confusion_matrix_baseline}
\end{figure}

\begin{figure}[htbp]
\centering
\includegraphics[width=0.9\textwidth]{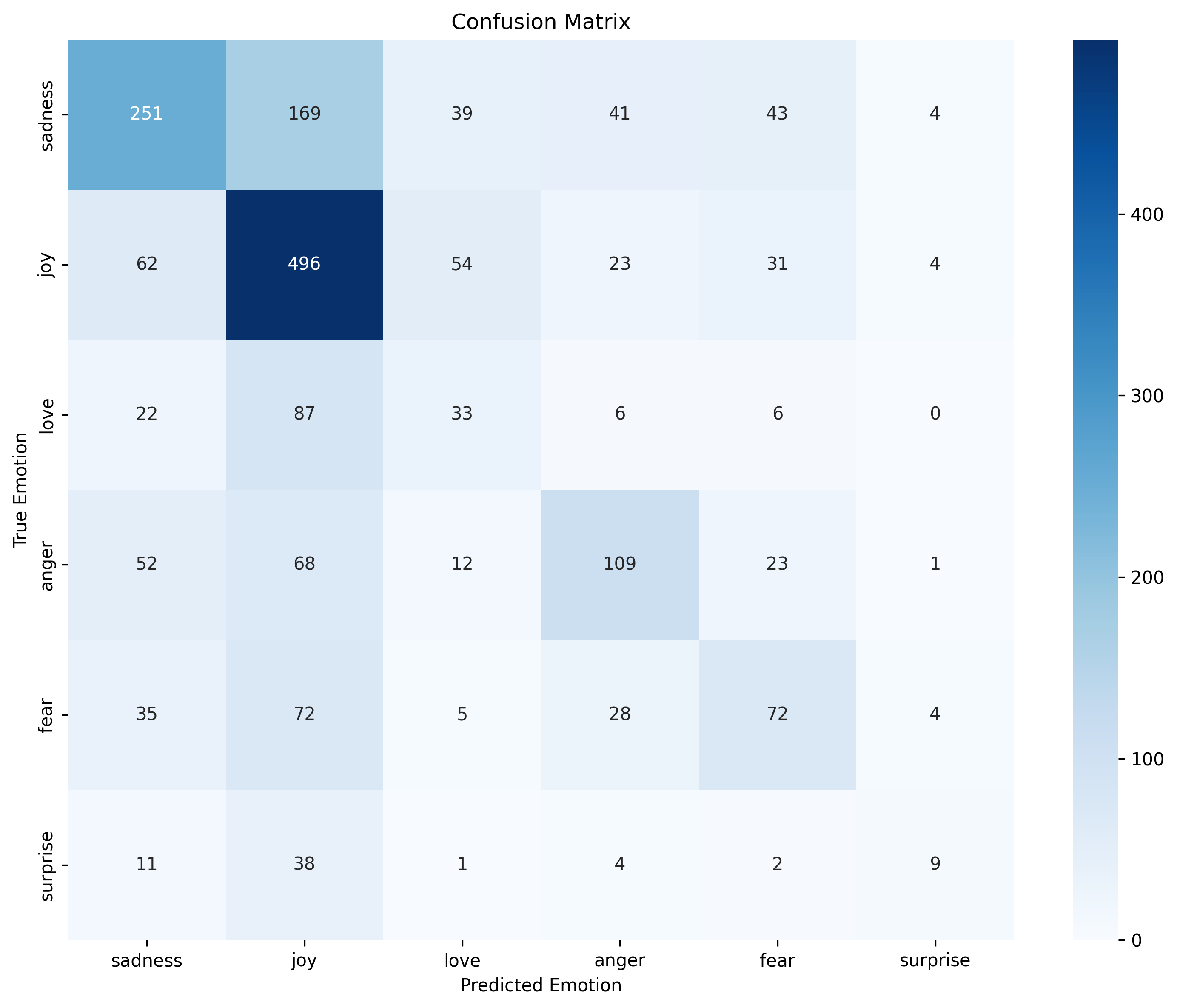}
\caption{Proposed Classifier (\texttt{Classifier\_Q->RA}): Confusion Matrix.}
\label{fig:confusion_matrix_reasoning}
\end{figure}

The confusion matrices (Figure \ref{fig:confusion_matrix_baseline} for Baseline, Figure \ref{fig:confusion_matrix_reasoning} for Proposed) reveal that the proposed model significantly reduces confusion between classes like sadness/joy and fear/joy. For instance, the baseline misclassified 169 sadness instances as joy, which our model reduced to 107. However, the matrices also confirm the collapse in performance for the "surprise" class, which is almost entirely misclassified as joy or sadness by the proposed model.

\section{Discussion}
\label{sec:discussion}
The results strongly suggest that training a model to explicitly generate reasoning as part of its output is a valuable strategy for improving classification performance. The statistically significant 8.7 percentage point accuracy improvement demonstrates that generating reasoning helps the model move beyond surface-level cues and develop a deeper understanding of the text.

\paragraph{Generalization of Reasoning.} A key finding is the successful transfer of reasoning ability from \texttt{Llama-R-Gen}, trained on logical problems (math, code, science), to the nuanced domain of emotion classification. This indicates that the fundamental patterns of constructing an argument or explanation learned from one domain can be effectively applied to another, even if the subject matter is completely different.

\paragraph{Quality of Generated Reasonings and Interpretability.} Our approach provides the dual benefit of enhanced performance and built-in interpretability. As shown in Table \ref{tab:rationale_examples}, the model often produces plausible explanations. However, the quality can vary. Assessing the "faithfulness" of these reasonings—whether they reflect the model's actual internal process—remains a core challenge in XAI. Future work should include human evaluation of the generated reasonings on metrics such as plausibility, faithfulness, and helpfulness in diagnosing model errors. The flawed reasoning in Case 2 (Table \ref{tab:rationale_examples}), where the model predicts 'fear' correctly but generates a justification for 'joy', highlights the complexity of this issue.

\paragraph{Analysis of Performance Degradation for the 'Surprise' Class.} The significant performance drop for the "surprise" class is a critical finding that highlights a limitation of our approach, particularly in the face of severe class imbalance. With only 66 test samples (3.3\% of the data), "surprise" is a minority class. This scarcity poses two problems: 1) The general-purpose \texttt{Llama-R-Gen} likely struggled to generate high-quality, specific reasonings for this rare and context-dependent emotion during data augmentation. 2) Training the downstream \texttt{Classifier\_Q->RA} on these few, potentially noisy (Text, Reasoning, Label) examples may have caused it to learn spurious correlations from the flawed reasonings, leading to a performance collapse. This underscores that our method's success is highly dependent on the quality of the generated reasonings, which can degrade for severely underrepresented classes.

\paragraph{Limitations.}
\begin{itemize}
    \item \textbf{Dependency on Initial Reasoning Generator:} The performance of the final classifier is inherently linked to the quality of the reasonings produced by \texttt{Llama-R-Gen}. Flawed or generic reasonings can introduce noise into the training process.
    \item \textbf{Computational Cost:} The two-stage process, involving fine-tuning and a large-scale offline generation step, is more computationally intensive than a single fine-tuning run.
    \item \textbf{Class Imbalance Sensitivity:} As shown with the "surprise" class, the method can be sensitive to severe class imbalance, where poor reasoning generation for minority classes can harm performance.
\end{itemize}

\paragraph{Broader Implications.} This work demonstrates a practical method for creating enriched "learning from explanations" datasets at scale. The successful transfer of reasoning capabilities and the single-model (Reasoning + Prediction) output architecture could be extended to other NLP tasks where intermediate steps are beneficial, such as natural language inference, question answering, and complex multi-label classification.

\section{Conclusion}
\label{sec:conclusion}
We introduced a two-stage reasoning-infused learning framework that significantly enhances text classification by training a generative model to produce explicit reasonings with its predictions. By fine-tuning a Llama-3.2-1B-Instruct model on a general reasoning dataset to augment emotion classification data, we successfully trained a downstream classifier that integrates reasoning into its learning process.

Our experiments on the \href{https://huggingface.co/datasets/dair-ai/emotion}{dair-ai/emotion} dataset demonstrated a statistically significant \textbf{8.7 percentage point} absolute improvement in accuracy over a strong baseline. This gain underscores the power of using LLMs to generate explanatory data, which serves as a rich supervisory signal, and confirms that models can generalize reasoning skills across disparate domains. While our approach showed strong performance on most emotion categories, its struggles with the highly imbalanced "surprise" class highlight the importance of reasoning quality and the challenges posed by data scarcity.

This study validates that learning to explain is a powerful mechanism for learning to predict. Future work will focus on improving reasoning generation for minority classes, developing methods to filter low-quality reasonings, and applying this framework to a broader range of NLP tasks.

\bibliographystyle{plainnat}
\bibliography{references}

\appendix
\section{Appendix: Hyperparameter Details}
\label{app:hyperparams}
This section provides detailed hyperparameters for model training.

\subsection{\texttt{Llama-R-Gen} Fine-tuning (Llama-3.2-1B-Instruct)}
\begin{table}[h!]
\centering
\caption{Hyperparameters for \texttt{Llama-R-Gen} fine-tuning.}
\label{tab:hyperparams_llama}
\begin{tabular}{lc}
\toprule
\textbf{Hyperparameter} & \textbf{Value} \\
\midrule
Base Model & Llama-3.2-1B-Instruct \\
Training Framework & Axolotl \\
GPU & NVIDIA A40 \\
Learning Rate & 2e-5 \\
Optimizer & \texttt{paged\_adamw\_8bit} \\
Learning Rate Scheduler & cosine \\
Warmup Steps & 100 \\
Weight Decay & 0.0 \\
Gradient Accumulation Steps & 8 \\
Micro Batch Size (per device) & 1 \\
Effective Batch Size & 8 \\
Num Epochs & 1 \\
Max Sequence Length & 16384 tokens \\
Sample Packing & True \\
Pad to Sequence Length & True \\
BF16 & auto \\
TF32 & False \\
Gradient Checkpointing & True (\texttt{use\_reentrant=false}) \\
Logging Steps & 1 \\
Flash Attention & True \\
Eval per Epoch & 2 \\
Saves per Epoch & 1 \\
Special Tokens & \texttt{pad\_token: <|end\_of\_text|>} \\
\bottomrule
\end{tabular}
\end{table}

\subsection{Downstream Generative Classifier Fine-tuning (Llama-3.2-1B-Instruct)}
\begin{table}[h!]
\centering
\caption{Hyperparameters for downstream generative classifier (Llama-3.2-1B-Instruct) fine-tuning.}
\label{tab:hyperparams_classifier}
\begin{tabular}{lc}
\toprule
\textbf{Hyperparameter} & \textbf{Value} \\
\midrule
Base Model & Llama-3.2-1B-Instruct \\
Training Framework & Axolotl \\
GPU & NVIDIA A40 \\
Learning Rate & 2e-5 \\
Optimizer & \texttt{paged\_adamw\_8bit} \\
Learning Rate Scheduler & cosine \\
Warmup Steps & 10 \\
Weight Decay & 0.0 \\
Gradient Accumulation Steps & 8 \\
Micro Batch Size (per device) & 2 \\
Effective Batch Size & 16 \\
Num Epochs & 1 \\
Max Sequence Length & 8192 tokens \\
Sample Packing & True \\
Pad to Sequence Length & True \\
BF16 & auto \\
TF32 & False \\
Gradient Checkpointing & True (\texttt{use\_reentrant=false}) \\
Logging Steps & 1 \\
Flash Attention & True \\
Eval per Epoch & 2 \\
Saves per Epoch & 1 \\
Special Tokens & \texttt{pad\_token: <|end\_of\_text|>} \\
\bottomrule
\end{tabular}
\end{table}

\end{document}